%% file: base.tex
\documentclass[preprint,12pt,authoryear]{elsarticle}

\usepackage{amssymb}
\usepackage{svg}
\usepackage{fixfoot}
\usepackage{times}
\usepackage{latexsym}
\usepackage{multirow}
\usepackage{tikz}
\usepackage{graphicx}
\usepackage{caption}
\usepackage{sidecap}
\usepackage{subcaption}
\usepackage{array}
\usepackage{multirow}
\usepackage{amsmath,amsfonts,bm}
\usepackage{microtype}
\usepackage{booktabs}
\usetikzlibrary{bayesnet}
\usetikzlibrary{arrows}
\usetikzlibrary{fit,positioning}
\usetikzlibrary{arrows, automata, positioning, quotes}

\input{math_commands.tex}

\journal{Expert Systems With Applications}

\begin{document}

\def\modelname{\emph{KitchenScale}}
\def\typeone{\textbf{Type 1}~}
\def\typetwo{\textbf{Type 2}~}

\begin{titlepage}
\begin{center}
\vspace*{1cm}

\textbf{ \large \modelname : Learning to Predict Ingredient Quantities from Recipe Contexts}

\vspace{1.5cm}

Donghee Choi$^{a}$ (choidonghee@korea.ac.kr), Mogan Gim$^a$ (akim@korea.ac.kr), Samy Badreddine$^c$ (samy.badreddine@sony.com), \\
Hajung Kim$^a$ (hajungk@korea.ac.kr),  Donghyeon Park$^b$ (parkdh@sejong.ac.kr), Jaewoo Kang $^a$ (kangj@korea.ac.kr) \\

\hspace{10pt}

\begin{flushleft}
\small  
$^a$ Korea University, Seoul, Korea \\
$^b$ Sejong University, Seoul, Korea \\
$^c$ Sony AI, Tokyo, Japan

\vspace{1cm}
\textbf{Corresponding Authors:} \\
Jaewoo Kang \\
Korea University, Seoul, Korea \\
Tel: +82 (02) 3290-4840 \\
Email: kangj@korea.ac.kr \\
\vspace{0.5cm}
Donghyeon Park \\
Sejong University, Seoul, Korea \\
Tel: +82 (02) 3408-1946 \\
Email: parkdh@sejong.ac.kr \\

\end{flushleft}        
\end{center}
\end{titlepage}

\begin{frontmatter}

\title{\modelname : Learning to Predict Ingredient Quantities from Recipe Contexts }

\author[korea_aff]{Donghee Choi}
\ead{choidonghee@korea.ac.kr}

\author[korea_aff]{Mogan Gim}
\ead{akim@korea.ac.kr}

\author[sony_aff]{Samy Badreddine}
\ead{samy.badreddine@sony.com}

\author[korea_aff]{Hajung Kim}
\ead{hajungk@korea.ac.kr}

\author[sejong_aff]{Donghyeon Park\corref{cor1}}
\ead{parkdh@sejong.ac.kr}

\author[korea_aff]{Jaewoo Kang\corref{cor1}}
\ead{kangj@korea.ac.kr}

\address[korea_aff]{%
    Korea University,
    Seoul,
	Korea
}
\address[sejong_aff]{%
    Sejong University,
    Seoul,
	Korea
}
\address[sony_aff]{%
    Sony AI,
    Tokyo,
	Japan
}

\cortext[cor1]{Corresponding author}

\begin{abstract}
Determining proper quantities for ingredients is an essential part of cooking practice from the perspective of enriching tastiness and promoting healthiness.
We introduce \modelname, a fine-tuned Pre-trained Language Model (PLM) that predicts a target ingredient's quantity and measurement unit given its recipe context.
To effectively train our \modelname~model, we formulate an ingredient quantity prediction task that consists of three sub-tasks which are ingredient measurement type classification, unit classification, and quantity regression task.
Furthermore, we utilized transfer learning of cooking knowledge from recipe texts to PLMs. 
We adopted the Discrete Latent Exponent (DExp) method to cope with high variance of numerical scales in recipe corpora.
Experiments with our newly constructed dataset and recommendation examples demonstrate \modelname’s understanding of various recipe contexts and generalizability in predicting ingredient quantities.
We implemented a web application for \modelname~to demonstrate its functionality in recommending ingredient quantities expressed in numerals (e.g., 2) with units (e.g., ounce).
\end{abstract}



\end{frontmatter}


\begin{figure}
\centering
	\includegraphics[width=0.5\textwidth]{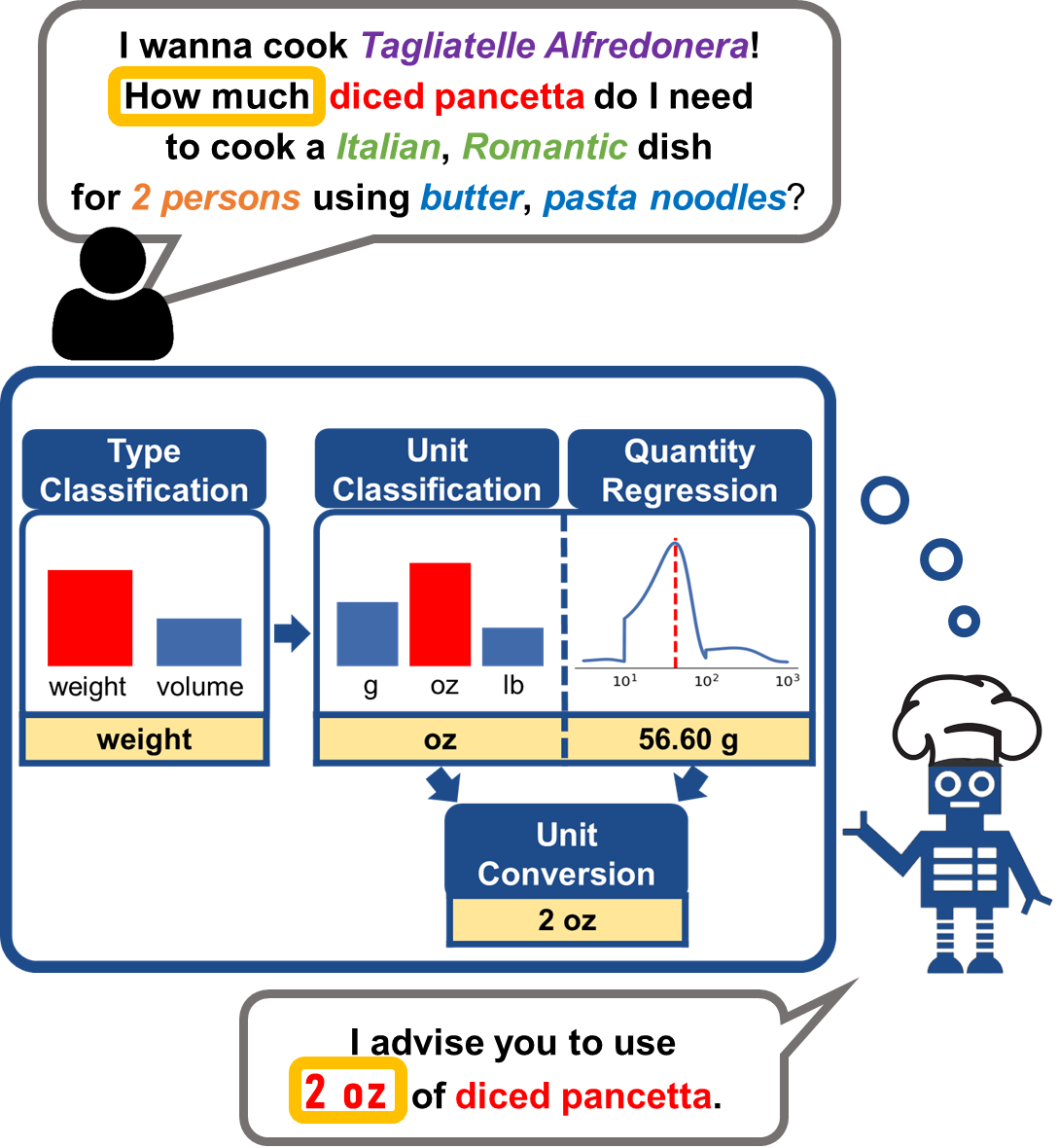}
	\caption{Overview of \modelname. The Ingredient Quantity Prediction task consists of 1) measurement type classification, 2) unit classification, and 3) quantity regression.
        The tasks are further explained in Section \ref{sec:task}.}
	\label{fig:task-overview}
\end{figure}

\section{Introduction}

Brainstorming new recipe ideas entails a series of complicated decision processes, which include determining reasonable ingredient quantities.
The amount of ingredients used in certain dishes plays a key role in enriching tastiness (flavor)~\citep{dalton2000merging} and promoting healthiness (nutrition)~\citep{dubois2012super, barrett1991nutrition}.
Recently, several data-driven approaches have leveraged large-scale recipe corpora such as Recipe1M~\citep{marin2019recipe1m+} to formulate various recipe-related downstream tasks such as recipe generation~\citep{h2020recipegpt,bien2020recipenlg}, recipe retrieval~\citep{li2020reciptor} and ingredient pairing~\citep{park2019kitchenette}.
However, such approaches seldom feature determining reasonable ingredient quantities though it is one of the important aspects in cooking knowledge~\citep{li2021deeprecipes}.
Some previous works have attempted to address such issues, but either suggest ingredient compositions expressed in ratios~\citep{li2021share,li2021deeprecipes} or treat numerals and their units as text~\citep{h2020recipegpt,bien2020recipenlg}.
This raises the necessity of developing a data-driven model capable of learning ingredient quantity prediction tasks.

Learning to predict ingredient quantities demands sophisticated cooking knowledge which is essentially related to ingredient quantity prediction.
PLMs which have been applied to various downstream tasks began to gain recognition for their acquisition of numeracy, an ability to understand the numerical aspects of given textual data.
Previous works have fine-tuned these models on a specific downstream task called numerical probing~\citep{wallace2019probe_numeracy,lin2020numersense, berg2020empirical}.
Adopting the above methodology, one may imagine to pre-train an existing language model on a large number of recipes with the objective of acquiring cooking knowledge and learning the ingredient quantity prediction task.

Several domain-specific issues should be addressed prior to formulating an ingredient quantity prediction task involving PLMs.
Contrary to general textual documents, cooking recipes are structured texts with multifaceted attributes that determine the overall recipe semantics~\citep{kiddon2016checklist_recipe_generation,batra2020recipedb}.
Such semantics would be beneficial for PLMs to learn cooking knowledge and accurately predict ingredient quantities.
Also, most ingredients are expressed in different measurement types (volume, weight) and units (ounces, cups) depending on the recipe context.
For example, \textit{crushed} tomatoes are measured in ounces while \textit{juiced} ones are measured in cups.
Furthermore, some ingredient quantities used in recipes have multi-modal distributional characteristics and high variance in numerical scales.
These properties are detrimental to traditional regression models which are mostly trained on mean squared error (MSE) loss criterion.

In this work, we introduce \modelname, a data-driven model that performs the ingredient quantity prediction task given recipe context as input.
We propose to adapt transfer learning from a large-scale corpus containing recipe texts to PLMs and utilize each of their multi-faceted attributes for a better understanding of recipe semantics.
To overcome the domain-specific challenges presented in the recipe corpus, we break the ingredient quantity prediction task into three sub-tasks which are 1) ingredient measurement type classification (volume, weight), 2) unit classification (ounces, cups, etc.), and 3) quantity regression.
To perform these formulated tasks, we constructed a dataset based on the 507,834 recipes from various data sources including RecipeDB~\citep{batra2020recipedb} and Reciptor~\citep{li2020reciptor}.

Our \modelname~model consists of three modules for each of our proposed sub-tasks that comprise the whole ingredient quantity prediction task.
Each module is equipped with a PLM that learns the contextual features from structured recipes.
The ingredient quantity regressor utilizes Discrete Latent Exponent~(DExp)~\citep{berg2020empirical} instead of the MSE-based regression method.
The DExp approach helps \modelname~cope with the multi-modal distribution of ingredient quantities and varying numerical scales.
As shown in Figure~\ref{fig:task-overview}, the ingredient type classification module in \modelname~predicts the \textit{measurement type} for the target ingredient in a given recipe context.
The predicted type (\textit{weight}) is used to augment the input for the other modules where unit classification and quantity regression is performed individually.
The two predictions (\textit{oz}, \textit{56.60 g}) are then converted into a final output with its predicted measurement unit (\textit{2 oz}).
Overall, \modelname~is given target ingredient with recipe context as input and recommends its reasonable quantity expressed in numeral + unit format as output which is illustrated in Figure~\ref{fig:task-overview}.

Quantitative experimental results demonstrate \modelname's robustness in predicting ingredient quantities.
Recommendation examples obtained from \modelname~demonstrate its understanding of various recipe contexts using different measurement types and units for the target ingredient.

Our contributions in this work are as follows,
\begin{enumerate}
	\item We introduce an ingredient quantity prediction task consisting in  three sub-tasks which are ingredient measurement type classification, unit classification, and quantity regression.
	~\footnote{https://github.com/dmis-lab/KitchenScale}
	\item We developed \modelname~which performs the ingredient quantity prediction task by adopting transfer learning of cooking knowledge from recipes to PLMs.
        \item To demonstrate the practicality of our approach, we converted an existing recipe dataset by extracting and refining large-scale textual recipes into a task of predicting ingredient quantities. 
	\item Experimental results demonstrate \modelname's robustness in predicting ingredient quantities which are attributed to the DExp approach for quantity regression task.
	\item We implemented a web-based application for \modelname~to demonstrate its functionality of recommending ingredient quantities.
	\footnote{http://kitchenscale.korea.ac.kr/}
\end{enumerate}

\section{Related Works}

\subsubsection*{Numerical Reasoning in Language Modeling}
While PLM has been widely used in general natural language processing tasks~\citep{devlin2019bert}, handling numerals has recently started to gain focus in the NLP domain~\citep{thawani2021nlp_number_survey}.
Various tasks including probing numerical knowledge~\citep{lin2020numersense, yamane2020measuring,elazar2019doq} and probing numerical embedding ~\citep{wallace2019probe_numeracy} showed the possibility of PLMs achieving numeracy.
While PLMs have been known to learn the scalar magnitude of objects~\citep{zhang2020language_capture_scale}, some domain-specific datasets may pose challenges to its learning on regression tasks due to their distribution characteristics.
Previous work tried a traditional linear regression architecture with simple scalar value estimation~\citep{wallace2019probe_numeracy}.
However, likelihood-based regression strategies have been proven to deal with numbers having high variance in their magnitudes~\citep{spithourakis2018numeracy}.
Recently, DExp method has been introduced and reported to achieve state-of-the-art performance compared to other likelihood functions-based regression approaches such as Gaussian Mixture Model, Transformed Laplace, and LogLaplace~\citep{berg2020empirical}.

For our \modelname~model, we adopted the DExp approach in our ingredient quantity regression task as it has shown success in previous works featuring numeral values with highly varying scales~\citep{berg2020empirical}.

\subsubsection*{Applications with Food Recipes}
The following approaches related to recipes and food ingredients have been proposed in the food computing domain.
Recipe embedding for downstream~\citep{li2020reciptor}, ingredients recommendation~\citep{park2019kitchenette, kim2021recipebowl, gim2022recipemind}, recipe recommendation~\citep{elsweiler2017food_biases_recommendation} and other recipe related tasks including recipe recommendation and recipe retrieval~\citep{lin2021subspace_crossmodal_retrieval_recipe_image} has been suggested.
Since the advent of Generative Pre-trained Transformer (GPT)~\citep{brown2020gpt_original}, recipe generation using fine-tuning language model via a large recipe corpus has gained attention~\citep{h2020recipegpt, bien2020recipenlg}.
However, such generative models rely on softmax classifiers that treat numerals as if they were words within the vocabulary corpus.  
These methods cannot handle the numerical values as regression tasks because measuring the difference between large and small numbers is often underestimated.

The Recipe1M dataset~\citep{marin2019recipe1m+} is a widely-used multi-modal dataset that contains over one million cooking recipes and 13 million food images. 
Many previous works~\citep{li2020reciptor,park2019kitchenette, kim2021recipebowl, gim2022recipemind, li2021deeprecipes, li2021share} on recipe mining have relied on this dataset, leveraging the vast amounts of user-generated recipes contained within it. 
In an effort to further enhance the dataset, RecipeDB~\citep{batra2020recipedb} and Reciptor~\citep{li2020reciptor} introduced more structured data elements such as ingredient phrases, units, servings, and tags using the Recipe1M+ dataset. 
In our work, we leveraged both the RecipeDB and Reciptor datasets by merging them to create a more comprehensive and structured dataset for our experiments.

Recently, DeepRecipe~\citep{li2021deeprecipes} and SHARE~\citep{li2021share} were proposed to predict ingredient compositions expressed in ratios given textual recipe context.
One example proposed in DeepRecipe is that given title "Hamburger", and other ingredients "salt, pepper, bread", the question is "how much is the proper ratio for meat?".
The answer might be 0.0017 which is calculated using the amount of the target ingredient dividing the total sum of the other ingredient's amount.
Another example for SHARE is that given two ingredients "salt, pepper" for a given title "hamburger", the question is predicting the proper ratio (e.g., salt 2: pepper 1) between them.
However, these approaches are limited to predicting ingredient compositions rather than \textit{exact quantities} of ingredients used in recipes.

In this work, we employed PLMs to enhance understanding of recipe contexts and formulated supervised learning tasks including \textit{ingredient quantity prediction} to achieve food numeracy beneficial to various recipe-related applications including recipe generation.

\section{Ingredient Quantity Prediction Task}
\label{sec:task}
Prior to introducing our tasks, we give the following definitions for the semantic elements in a cooking recipe,
\begin{itemize}
    \item Recipe $R$:
    A recipe is a description of various food-related semantics that help the user cook a resulting dish.
    The semantics that comprises the recipe is the Recipe Title $e$, Ingredients $I$, Descriptive Tags $B$, and Number of Servings $s$.
    \item Ingredients $I$: A set of $n$ fundamental elements in a recipe that comprise and determine the overall flavor and texture of the resulting dish. We denote each element of $I$ as $I = \{ i_{0}, i_{1} ... i_{n-1}\}$ where $|I|=n$.
    \item Target Ingredient $i_t$: A single ingredient among $I$ designated as target for the numeracy probing tasks ($i_t\in I$).
    \item Other Ingredients $I_{o}$: A set of ingredients where $|I_{o}\cap\{i_{t}\}|=0$ and $I_{o}\cup\{i_{t}\}=I$.
    \item Title $e$: The title of the recipe which briefly describes the resulting dish.
    \item Tags $B$: A set of $m$ tags that provide specific information about the dish such as cuisine region, cooking time, dish category, etc. We denote each element of $B$ as $B = \{ b_{0}, b_{1} ... b_{m-1}\}$ where $|B|=m$.
    \item Number of Servings $s$: A scalar value that can be used to adjust the quantity of ingredients used when cooking the same dish.
    \item Recipe Context Query Composition $C$: A recipe context query consists of the textual elements extracted from the semantics in source recipe $R$. Each numeracy probing task uses $C = \{i_{t},I_{o},B,e,s\}$ as input.
\end{itemize}

We then make the following definitions for the textual elements of the target ingredient $i_{t}$ which are deeply related to our proposed task.
\begin{itemize}
    \item Descriptive Name $i_{t}^{DescName}$: The name of the target ingredient which includes descriptive words such as `diced' and `sliced'. Target ingredient names whose descriptive words were removed are denoted $i_{t}^{Name}$.
    \item Measurement Type $i_{t}^{d}$: An ingredient property that determines the method of measurement. In this work, we use two labels \textit{Volume} and \textit{Weight}.
    \item Measurement Unit $i_{t}^{u}$: A standard quantity to express the target ingredient's physical ingredient. In this work, we use 14 labels for $U \in \{
    cup, tablespoon, \\ teaspoon, lb, ounce, g, ml, pinch, dash, kg, pint, quart, drop, gallon\}$.
    \item Quantity $i_{t}^{q}$: A continuous value that represents the amount of target ingredient used in its recipe.
\end{itemize}

Our proposed food numeracy probing tasks are as follows,
\begin{problem}[Ingredient Quantity Prediction Task]
\label{problem:food_numeracy}
Given a recipe context $C$ which must include the target ingredient $i_t$, we define the objective of the \textbf{Ingredient Quantity Prediction Task} as predicting the target ingredient's measurement type $d$, unit $u$ and normalized quantity $q$.
\end{problem}
To solve this problem, we break it down into three different numeracy tasks.

\begin{task}[Ingredient Measurement Type Classification]
    \label{task:measurement-type-prediction}
	Given the recipe context $C$ and the target ingredient, \modelname's objective is to predict its correct measurement type $d$ through modeling $P(d|C)$, where $d \in \{\textit{Weight}, \textit{Volume}\}$,  $C = \{i_t^{DescName}, I_{o}, e, B, s\}$.
\end{task}

\begin{task}[Ingredient Measurement Unit Classification]
    \label{task:measurement-unit-prediction}
	Given the recipe context $C$, measurement type $i_t^{d}$,  and the target ingredient, \modelname's objective is to predict its correct measurement unit $u$ through modeling $P(u|C, i_t^{d})$, where $u \in U$, $C = \{i_a^{DescName}, I_{o}, e, B, s\}$, $i_{t}^{d} \in \{\textit{Weight}, \textit{Volume}\} $.
\end{task}

\begin{task}[Ingredient Quantity Regression]
    \label{task:quantity-prediction}
	Given the recipe context $C$, measurement type $i_t^{d}$ and the target ingredient, \modelname's objective is to estimate its exact normalized quantity $q$ through modeling $P(q|C, i_t^{d})$, where $q\in\mathbb{R}$,
	$C = \{i_{a}^{DescName}, I_{o}, e, B, s\}$, $i_{t}^{d} \in \{\textit{Weight}, \textit{Volume}\}$.
\end{task}

\section{Dataset}

\subsection{Dataset Construction}
\label{subsection:data-construction}

\begin{table}[]
\centering
    \scalebox{1}{
	\begin{tabular}{l r r r r}
		\toprule
		                 & \textbf{train} & \textbf{valid} & \textbf{test} \\
		\hline
		$\#$ of instances. & 78,984         & 9,873          & 9,868         \\
		\hline
		\multicolumn{4}{l}{\textbf{Statistics for  target ingredient quantity}} \\
		mean             & 185.93         & 187.29         & 182.36        \\
		std              & 384.88         & 383.83         & 385.41        \\
		\hline
		min              & 0.05           & 0.05           & 0.05          \\
		max              & 30,283.28      & 15,141.64      & 11,356.32     \\
		\bottomrule
	\end{tabular}
	}
	\caption{Dataset Statistics: Numbers in the first row are counts of instances. Details are in Section ~\ref{subsection:data-construction}.}
	\label{table:dataset_stats}
\end{table}

We built the dataset for \modelname's ingredient quantity prediction task using the recipes and their attributes from RecipeDB\citep{batra2020recipedb} and Reciptor\citep{li2020reciptor} dataset.
Both of these original datasets were merged into one dataset including recipe information, tags where the statistics are shown in Table~\ref{table:dataset_stats}.
The merged number of recipes is 101,573.

We then randomly selected a target ingredient from the list of used ingredients for each recipe.
The numeral information associated with the target ingredient is masked accordingly to our ingredient quantity prediction tasks.
For each target ingredient in a recipe, the masking process is done by matching the textual representation in numeric values and units in RecipeDB and the original ingredient text.

While a wide variety of measurement units were used to describe each ingredient's quantity information, we only used 74 units whose occurrence exceeded 100 counts in our dataset.
We preprocessed the units by disambiguating the abbreviations (e.g., converting pounds to lb) and normalizing plural textual representations into singular ones (e.g., tablespoons to tablespoon).
According to the International System of Units (SI) \citep{mechtly1964international_system_of_units}, we use 14 measurement units and 2 (Volume, Weight) measurement types.
Note that we used \emph{ml} for Volume and \emph{g} for Weight as base units for numeral conversion.
We discarded data instances where the measurement unit of the target ingredient is either unknown or not among the selected 14 units.

Finally, we converted the numeric quantity of each recipe's target ingredient into a normalized float value by following the procedures.
We converted the fraction-based numerals to their corresponding decimal values (e.g., `1 1/2' to 1.5) to formulate a quantity prediction problem as a regression problem.
We used the standardized conversion to normalize that decimal value for each measurement type of its unit.

As a result, we obtained 98,725 data instances where each data instance is defined as $ K = \{d, u, q, c\}$ where $c = \{i_{t}^{DescName}, I_{o}, e, B, s\}$.
The data instances in the \modelname~dataset were split into $\train : \valid : \test = 8:1:1 $.
Table~\ref{table:dataset_stats} shows the statistics of quantities of target ingredient.


\section{KitchenScale}
\label{sec:model}

\begin{figure}
\centering
\begin{subfigure}[b]{0.38\textwidth}
	\includegraphics[width=\textwidth]{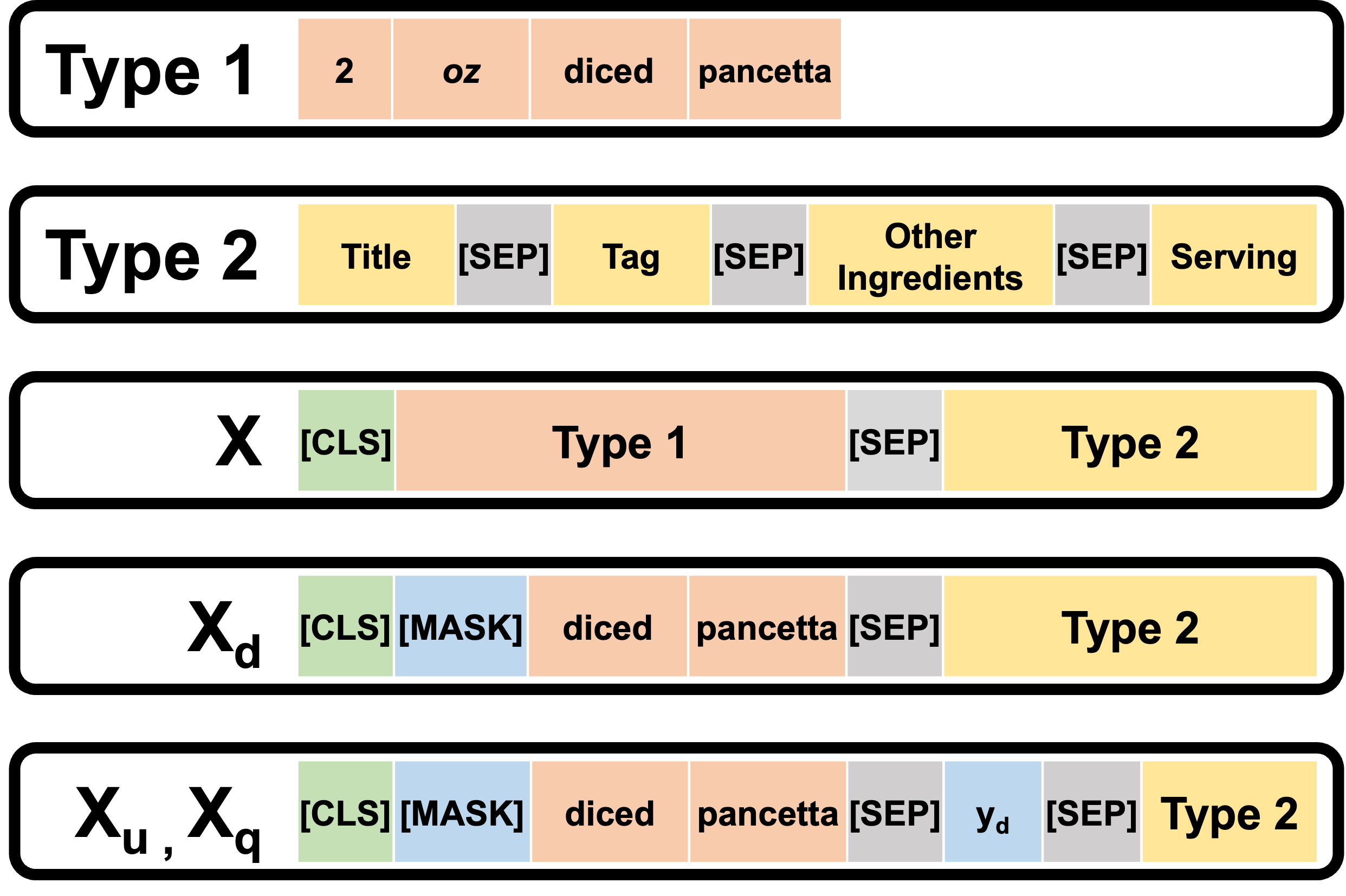}
	\caption{Input composition of \modelname}
	\label{fig:model-overview:input}
\end{subfigure}
\hspace{1cm}
\begin{subfigure}[b]{0.4\textwidth}
    \includegraphics[width=\textwidth]{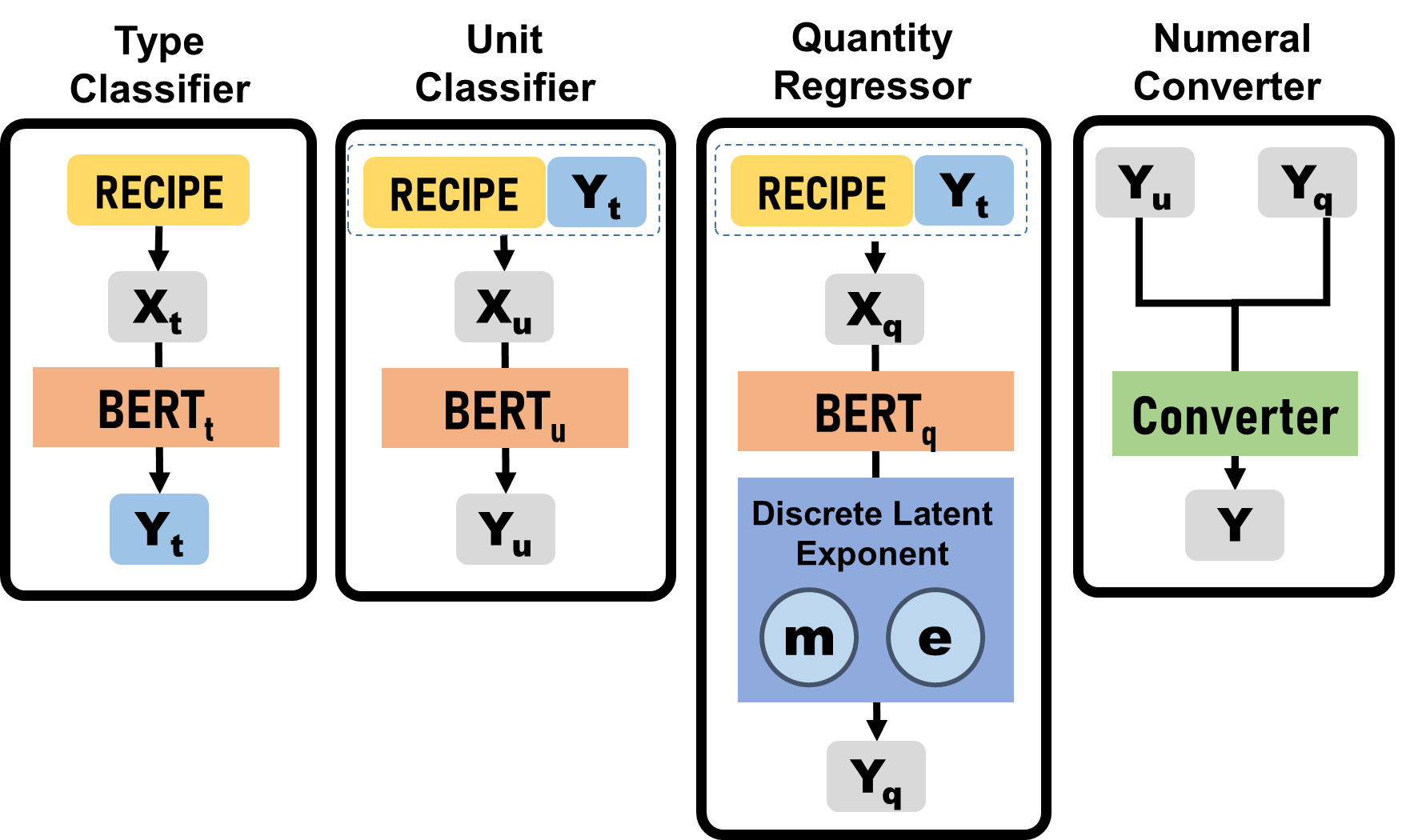}
	\caption{Architecture of \modelname }
	\label{fig:model-overview:architecture}
\end{subfigure}
\caption{Model architecture of \modelname~described in Section~\ref{sec:model}  }
\label{fig:model-overview}
\end{figure}

\subsection{Overview}
\label{subsection:model-overview}
\modelname~takes recipe query context containing multi-type semantics and a masked target ingredient as input to predict its numeral elements.
The numeral elements are measurement type, unit, and quantity used for cooking the dish in the recipe.
Our model architecture has three different prediction modules which are the Ingredient Measurement Type, Unit, and Quantity Predictor.
Each prediction module has its own input query composition and consists of a recipe context encoder and a prediction layer for its corresponding numeracy task.

All prediction modules use two types of textual elements extracted from an actual recipe to comprise an input recipe query for \modelname.
We denote textual elements related to recipe attributes as \typetwo and target ingredient as \typeone.
\typetwo textual elements used in all prediction modules concatenated by a special token `[SEP]' are the followings,
\begin{itemize}
    \item Title of the Recipe
    \item Descriptive Tags of the Recipe where multiple tags are concatenated by another special token `[SEP2]'.
    \item Other Ingredients used in the Recipe which are also concatenated by another special token `[SEP2]'.
    \item Number of Servings where fractions are converted to decimals.
\end{itemize}

For each prediction module, some of the task-specific \typeone textual elements concatenated by `[SEP]' are replaced by a special token `[MASK]'.
The two different types of textual elements are then concatenated by a `[SEP]' token and are deemed as recipe input query composition $X_{f}$ where $f$ refers to a numeracy task.
We use `[SEP2]' for concatenating sub-structure elements ( between tags and other ingredients ) to distinguish with top-level elements which are described above ( title, tags, .. ) and some works used `[SEP2]' for giving other structural information ~\citep{jin2019learning_sep2, chakrabarty2021s_sep2, lin2021pretrained_sep2}.

Further explanation for recipe input query composition regarding \typeone textual elements is given later.

For user-friendly purposes, we further added the Final Numeral Converter module that uses the outputs from Ingredient Measurement Unit and Quantity Predictor.
The Final Numeral Converter module re-converts the predicted value to an actual quantified value of the target masked ingredient by using the measurement unit.
Figure~\ref{fig:model-overview} shows the outline of our models and input composition method.

\subsection{Ingredient Measurement Type Classifier}
The \typeone textual elements in the input query for Ingredient Measurement Type Predictor module are the followings,
\begin{itemize}
    \item Descriptive Name of the Target Ingredient.
    \item Measurement Unit of the Target Ingredient masked by a special token `[MASK]'.
    \item Quantity of the used Target Ingredient also masked by a special token `[MASK]'.
\end{itemize}
which gives recipe input query $X_{d}$ when concatenated with the \typetwo textual elements.
Note that we used only one special token `[MASK]' to replace the multiple masked textual elements in our input query composition.

Next, the recipe context encoder in Ingredient Measurement Type Predictor encodes $X_{d}$ into a contextual representation $H_{d}$ using BERT as the encoder model.
For obtaining $H_{d}$, the final hidden state corresponding to the first input token `[CLS]' is forwarded to the prediction layer as suggested in \citep{devlin2019bert}.
We imported the BERT model from Hugging Face~\citep{wolf-2020-huggingface} where the pre-trained model is named `bert-based-uncased' and all other Predictor modules employ the same encoding scheme with differently trained task-specific BERT models.
Therefore, while a BERT model has 112M trainable parameters, the recipe context encoders in \modelname~have a total of 336M.

The contextual representation $H_d$ is fed to a classification layer that predicts a score for each class label (e.g., Volume, Weight).
In sum, the ingredient measurement type prediction task performed by this module taking $X_d$ as input is mathematically expressed as,
\begin{align}
	\label{eq:encoder_imtp}
	H_d = BERT_{d}{(X_d)},  y_d = Linear_{d}(H_d)
\end{align}
where $H_d \in \mathbb{R}^{768}$ is a $768$-dimensional contextual representation,
and $y_d$ has two score predictions for each class label.

\subsection{Ingredient Measurement Unit Classifier}
The \typeone textual elements in the input query for Ingredient Measurement Unit Predictor module are the followings,
\begin{itemize}
    \item Descriptive Name of the Target Ingredient.
    \item Measurement Unit of the Target Ingredient masked by a special token `[MASK]'.
    \item Quantity of the used Target Ingredient also masked by a special token `[MASK]'.
    \item Measurement Type of the Target Ingredient (\textit{Volume}, \textit{Weight}).
\end{itemize}
which gives recipe input query $X_{u}$ when concatenated with the \typetwo textual elements.

Similar to the Measurement Type Predictor module, the unit prediction task performed by this module taking $X_u$ as input is mathematically expressed as,
\begin{align}
	\label{eq:encoder_imup}
	H_u = BERT_{u}{(X_u)}, y_u = Linear_{u}(H_u)
\end{align}
where $H_u \in \mathbb{R}^{768}$ is a $768$-dimensional contextual representation,
$y_u$ has 14 score predictions for each class label.

\subsection{Ingredient Quantity Regressor}
\label{sec:quantity-predictor}
The \typeone textual elements in the input query for Ingredient Quantity Predictor module are the followings,
\begin{itemize}
    \item Descriptive Name of the Target Ingredient.
    \item Measurement Unit of the Target Ingredient masked by a special token `[MASK]'.
    \item Measurement Type of the Target Ingredient.
\end{itemize}
which gives recipe input query $X_{q}$ when concatenated with the \typetwo textual elements.

According to Table~\ref{table:dataset_stats} some target ingredients in our \modelname~dataset have multimodal distributions of quantity values with a high variance which poses a challenge to training \modelname's regression task.
We utilized DExp component for implementing the Ingredient Quantity Predictor module which consists of sub-components exponent ($\mathbf{e}$) and mantissa ($\mathbf{m}$)~\citep{berg2020empirical}.
The sub-components $\mathbf{e}$ and $\mathbf{m}$ consist of a 3-layered MLP followed by a sampling scheme from a multinomial distribution and truncated Gaussian.

The following equations taking $H_q$ describe how the DExp approach outputs the ingredient quantities using the above variables.
\begin{align}
	\label{eq:encoder_iqp_exp_gener}
	H_q &= BERT_{q}{(X_q)} \\
	e_q &\sim Mult(\pi_{\theta}(H_q)) \\
	y_q &\sim N_{trunc[0.1,1]}(\mu_\theta(H_q, e_q), 0.05)
\end{align}

$\pi_{\theta}$ is for predicting exponent component, and $\mu_{\theta}$ is for mantissa part.
And the following equations describe the forward pass of the DExp regression layer.
\begin{align}
    \label{eq:forward_dexp}
    \pi_{\theta} &= Linear_{q}(H_{q})\\
    \mu_{\theta}(X, i) &= Scale(MLP_{q}(X))_{i} \\
	Scale(X) &= X * 0.9 + 0.1
\end{align}
where
$7$ is the dimension for exponent values, and $MLP_{q}$ composes 2 linear layers with sigmoid activation between them.
We determined this value based on the training set, which ranges from a minimum order of magnitude of $10^{-2}$ to a maximum of $10^4$.
For $\mu_{\theta}$, $i$ is used for selecting a value the $i$th index.

$\pi_{\theta}$ is a single linear layer for capturing the multinomial parameters of $P(e|X)$.
$\mu_\theta$ is a 3-layered MLP using sigmoid as non-linear activation for modeling the mean parameter of mantissa normal distribution.
$e$ is a sampled latent variable for emitting the mean $\mu$ of a normal distribution which uses a fixed standard deviation of 0.05.

The training objective for \modelname~using negative likelihood loss criteria is mathematically expressed as follows,
\begin{align}
	\label{eq:encoder_iqp_exp_train1}
	e^*(y_q) &= \lfloor {log_{10}{(y_q)}} \rfloor \\
    \begin{split}
	logP(y_q|X_q) &= log[P(e=e^*(y))  \\
	\qquad \qquad \qquad & \cdot \frac{1}{C} exp(-10( \frac{y}{10^{e^*(y) ) } } - \mu_\theta(H, e^*(y)))^2 )]
	\end{split}
\end{align}

where $C = |log(1/y)| $

For the test time, we use maximum likelihood value $\hat{y}_{q}$ as prediction values with $\hat{e}_{i}$ as most probable exponent value, expressed as follows,
\begin{align}
	\label{eq:encoder_iqp_exp_train2}
	\hat{e}_{i} &= \mbox{argmax}(\pi_{\theta}(H_q)) \\
	\hat{y}_{q} &= \mu_\theta(H_q,\hat{e}_{i})\cdot 10^{(\hat{e}_{i})}
\end{align}
where argmax is for selecting the index that has the biggest value among other exponent indices.

In terms of computational complexity, the main bottleneck of this model depends on the layers in BERT-based modules.
As stated in the original transformer paper~\citep{vaswani2017transformer}, each layer in the BERT-based module has $O(d \cdot n^2)$ complexity, where $d$ is the representation dimension and $n$ is the length of input.
So in \modelname, $d \in \{H_d, H_u\}$ is 768 mentioned in equation \ref{eq:encoder_imtp} and equation \ref{eq:encoder_imup}.
Also, as we use `bert-base' model, $n$ is 512 and the number of layers are 12.
But the transformer model gets input as a whole, the sequential complexity is $O(1)$ compared to sequential approaches like RNN and LSTM are $O(l)$ where $l$ is the sequence length.

\subsection{Final Numeral Converter}
Using the predicted $y_d$, $y_u$, and $y_q$ values, we merged them into a single user-friendly output $y$.
The quantity value $y_q$ is a normalized value using the measurement type  $y_d$, so we calculate the converted quantity $\frac{y_d} {f_{si}{(y_u)}}$ where $f_{si}$ is finding a factor for a given unit.
For example, if the predicted results are $y_d = amount$, $y_u = ounce$, $y_q = 109.774 $ , the converted result $y$ is \emph{3.875 ounces}.

For user-friendly purposes, we further added the Final Numeral Converter module to re-convert the predicted value to an actual quantified value of the target masked ingredient.
More discretized quantity expressions help the user's discrimination ability\citep{lembregts2019unit_quantity_discrete}.
So we make the following list contains the top 10 most frequent expressions for fractions $\{ \frac{1}{16}, \frac{1}{8}, \frac{1}{4}, \frac{3}{8}, \frac{1}{3}, \frac{1}{2}, \frac{5}{8}, \frac{2}{3}, \frac{3}{4},  \frac{7}{8}\}$.
Based on the list, we matched fraction part to a fraction.
For example, \emph{3.875 ounces} has \emph{0.875} as fraction part, we convert it into $\frac{7}{8}$.
The final conversion result of the predicted results are $y_d = amount$, $y_u = tablespoon$, $y_q = 38.82375 $ is $y = 3\frac{7}{8} tablespoons$

\section{Experiments}

\subsection{Model Training and Evaluation}
We conducted several experiments to evaluate and compare \modelname's performance in three different numeracy tasks with its baselines and input ablations.
The evaluation metrics for assessing \modelname's numeracy in Ingredient Measurement Type and Unit Classification Task are Accuracy (Acc).

For the Ingredient Quantity Regression Task, the evaluation metrics are primarily Mean Squared Error (MSE) and Mean Absolute Error (MAE).
As previously mentioned, the distribution of ingredient quantities in our dataset has high variances which render the above evaluation metrics sensitive to high magnitudes of prediction error.
The following evaluation metrics to compensate for this issue were additionally used in our experiments.

\begin{equation}
	\label{eq:metric:regression2}
	\begin{aligned}
		\text{\emph{LMAE}} &= \frac{1}{|\test|}\sum_{\test}|\log\num - \log\pred| \\
        \text{\emph{E-Acc}} &= \frac{1}{|\test|}\sum_{\test}=
		\left\{\begin{matrix}
			       1 & if \floor{\log\num} = \floor{\log\pred} \\
			       0 & otherwise
		       \end{matrix}\right.
	\end{aligned}
\end{equation}

\begin{itemize}
    \item \textbf{MAPE} : Mean Absolute Percentage Error. Originally used in DeepRecipes, this metric alleviates the sensitivity issue mentioned above~\citep{li2021deeprecipes}.
    \item \textbf{LMAE} : Log Mean Absolute Error. Originally used in DExp, this metric rewards \modelname's predicted values similar to actual values in terms of magnitude~\citep{berg2020empirical}.
    \item \textbf{E-Acc} : Exponent Accuracy. Also originally used in DExp, this metric was used to specifically evaluate \modelname's Exponent component~\citep{berg2020empirical}.
\end{itemize}

\subsection{Model Baseline and Input Ablations}
We conducted experiments to evaluate and compare our proposed \modelname's performance on Task~\ref{task:measurement-type-prediction},\ref{task:measurement-unit-prediction},\ref{task:quantity-prediction} with other baselines and input ablations.

We made the following ablations on \modelname's input textual representation and BERT configuration in the recipe context encoder.
Since fine-tuning the pre-trained weights of BERT imported from Hugging Face~\citep{wolf-2020-huggingface} is our default setting in \modelname, we denote it as \emph{BERT} for brevity.
\begin{itemize}

    \item \textbf{W2V + Bi-GRU or MLP}:
    We imported the word2vec (W2V) embeddings from the GLOVE-6B model~\citep{pennington2014glove} and replaced the BERT-based encoder with Bidirectional Gated Recurrent Units (Bi-GRU)~\citep{cho2014gru} or three Linear MLP layers.
    For MLP layers, the last layer of latent vectors are fully connected to the first of MLP layers.

    \item $\textbf{{BERT}}^{freeze}$:
    We included this ablation to quantitatively examine the benefits of fine-tuning the pre-trained weights of BERT~\citep{levine2022freeze}.

    \item $\textbf{{BERT}}^{init}$:
    For this setup, we randomly initialized BERT without using pre-triained weight.
    We included this ablation to see whether initializing the PLM with pre-trained weights imported from the `bert-based-uncased' model in Hugging Face benefits \modelname~in achieving numeracy.

\end{itemize}

In addition, we made the following ablations on \modelname's ingredient Quantity Regression module as described in Section~\ref{sec:quantity-predictor}.
\begin{itemize}
    \item $\textbf{MLP + L1Loss}$: We replaced the DExp regressor with an MLP followed by a feedforward layer which is mainly point-wise estimation in regression task and employed the L1 Loss training objective to facilitate stability of its training~\citep{berg2020empirical, thawani-etal-2021-numeracy, zhang2020language_capture_scale, spokoyny2021masked}.
    \item $\textbf{LogLP}$: 
    We use the log Laplace distribution to create probability density functions that are expressive in various scales~\citep{berg2020empirical}. 
    Forward pass is expressed as follows:
    \begin{align}
	\label{eq:encoder_loglp_gener}
	z &\sim Laplace(\mu_{\theta}(H),1) \\
        y &= g_\theta(z) = exp(z)
    \end{align}
\end{itemize}

Finally, we conducted ablations on \modelname's multi-type input recipe query composition by modifying the \typeone and/or \typetwo textual elements comprising it.
The default input query composition for \modelname~is denoted as $T_{DescName}+All$ for the Ingredient Measurement Type Classification task and $T_{DescName}+T_{Type}+All$ for both Ingredient Measurement Unit and Quantity Prediction task.
$T_{DescName}$ is the descriptive name of the target ingredient (same as $i_{t}^{DescName}$), $T_{Type}$ is its measurement type and $All$ refers to using all \typetwo textual elements as explained in Section 5.
\begin{itemize}
    \item $\textbf{Title, OtherIngs, Tags, Servings}$: This ablation uses only one of the \typetwo textual elements which are the recipe title, list of other ingredients ($OtherIngs$), list of descriptive tags and a number of servings. For example, $T_{DescName} + T_{Type} + Title$ means only the target ingredient's descriptive name, measurement type, and title of the corresponding recipe is included in the input query composition.
    \item ${\textbf{T}_{\bm{Name}}}$: Since most ingredient names are represented with descriptive words such as \textit{diced}, we performed an ablation using only the primary name of the target ingredient $i_{t}$. For instance, we replaced \textit{medium red bell pepper, diced} with \textit{red bell pepper} (same as $i_{t}^{Name}$).
    \item ${\textbf{T}_{\bm{PredType}}}$: We replaced the target ingredient's actual measurement type with its predicted label from \modelname's Ingredient Measurement Type Classifier module. Note that we deployed \modelname~in real-world applications using $T_{DescName}+T_{PredType}+All$ as input query composition.
\end{itemize}

\begin{table}[]
\centering
 \scalebox{1}{
    \renewcommand*{\arraystretch}{1.2}

	\begin{tabular}{l l r }
		\toprule
		                     & \textbf{Context}       & \textbf{Acc$\uparrow$}    \\
		\hline
		$W2V + BiGRU    $    & $T_{DescName}$         & 0.8329          \\
		$W2V + MLP      $    & $T_{DescName}$         & 0.8944          \\
		$BERT^{freeze}  $    & $T_{DescName}$         & 0.8348          \\
		$BERT^{init} $    & $T_{DescName}$         & 0.9100          \\
		$BERT $              & $T_{DescName}$         & 0.9186          \\
		\hline
		$W2V + BiGRU    $    & $T_{DescName} + All$   & 0.8329          \\
		$W2V + MLP      $    & $T_{DescName} + All$   & 0.8822          \\
		$BERT^{freeze} $     & $T_{DescName} + All$   & 0.8329          \\
		$BERT^{init}$     & $T_{DescName} + All$   & 0.8591          \\
		\hline
		\multicolumn{1}{c}{\textbf{\modelname~for Task \ref{task:measurement-type-prediction}}}               \\
		$BERT$               & $T_{DescName} + All$   & \textbf{0.9233} \\
		\bottomrule
	\end{tabular}
	}
	\caption{Measurement Type Classification Results for Task \ref{task:measurement-type-prediction}.
	$T_{DescName}$ means input is composed of only target ingredients text with quantity and unit parts are masked out.
	$All$ is about an input that has all of the recipe contexts defined in Task\ref{task:measurement-type-prediction}.
	}
	\label{table:measurement-type-prediction}
\end{table}

\begin{table}[]
\centering
    \renewcommand*{\arraystretch}{1.2}
    \scalebox{1}{
	\begin{tabular}{l l r}
		\toprule
		                            & \textbf{Context}                 & \textbf{Acc$\uparrow$}    \\
		\hline
		$W2V+BiGRU$                 & $T_{DescName} + T_{Type}$        & 0.3477          \\
		$W2V+MLP$                   & $T_{DescName} + T_{Type}$        & 0.6186          \\
		$BERT^{freeze} $            & $T_{DescName} + T_{Type}$        & 0.4042          \\
		$BERT^{init}$            & $T_{DescName} + T_{Type}$        & 0.7119          \\
		$BERT $                     & $T_{DescName} + T_{Type}$        & 0.7200          \\
		\hline
		$W2V+BiGRU$                 & $T_{DescName} + T_{Type} + All$  & 0.3477          \\
		$W2V+MLP$                   & $T_{DescName} + T_{Type} + All$  & 0.6531          \\
		$BERT^{freeze}$             & $T_{DescName} + T_{Type} + All$  & 0.3476          \\
		$BERT^{init}$            & $T_{DescName} + T_{Type} + All$  & 0.7072          \\
		\hline
		\multicolumn{1}{c}{\textbf{\modelname~for Task \ref{task:measurement-unit-prediction}}}                    \\
		$BERT$               & $T_{DescName} + T_{Type} + All$         & \textbf{0.7335} \\
		$BERT$               & $T_{DescName} + T_{PredType} + All$     & \textbf{0.6831} \\
		\bottomrule
	\end{tabular}
	}
	\caption{
	Measurement Unit Classification Results for Task \ref{task:measurement-unit-prediction}.
	$All$ is about an input that has all of the recipe contexts defined in Task\ref{task:measurement-unit-prediction}.
	$All_{TypePred}$ is using the same recipe context with $All$ except the dimension.
	$All_{TypePred}$ is using the best model that is used for Task~\ref{task:measurement-type-prediction} in Table~\ref{table:measurement-type-prediction} ( $BERT$ with $All$ )
	}
	\label{table:measurement-unit-prediction}
\end{table}

\begin{table}[]
\renewcommand*{\arraystretch}{1.1}
\begin{subfigure}[c]{\linewidth}
    \centering
    \scalebox{1}{
	\begin{tabular}{l r r r r}
		\toprule
		                              & \textbf{E-Acc$\uparrow$} & \textbf{LMAE$\downarrow$} & \textbf{MAPE$\downarrow$} & \textbf{MAE$\downarrow$} \\

        \hline
		\multicolumn{1}{c}{\textbf{Likelihood Model Ablation}}                                                                                      \\
            $W2V+MLP+L1Loss    $ & 0.2207 & 2.0712 & 1055.3033 & 6652.8691 \\ 
            $W2V+BiGRU+L1Loss    $ & 0.2207 & 2.2494 & 1245.5807 & 8001.6699 \\
            $BERT+MLP+L1Loss      $       & 0.2207                   & 2.2603                    & 1275.4724                 & 8206.8232                \\
            $W2V+MLP+LogLP    $ & 0.3389 &  0.6903  & 8.6111 & 164.8311 \\
            $W2V+BiGRU+LogLP    $ & 0.3389 &  0.6903  & 8.6385 & 164.8291 \\
            $BERT+MLP+LogLP    $ & 0.3389 &  0.8224  & 2.3670  & 172.6675 \\

		\hline
		\multicolumn{1}{c}{\textbf{Encoder Ablation}}                                                                                               \\
		$W2V+MLP+DExp  $          & 0.6677                   & 0.4364                    & 4.3531                    & 136.7550                  \\
		$W2V+BiGRU+DExp$          & 0.3957                   & 0.9434                    & 49.1616                    & 280.7921                  \\
		$BERT^{freeze}+DExp $     & 0.4020                   & 0.9117                    & 44.8252                   & 264.9828                 \\
		$BERT^{init}+DExp $ & 0.7097                   & 0.3712                    & 3.9536                    & 121.4156                 \\
		\hline

		\multicolumn{1}{c}{\textbf{\modelname~for Task \ref{task:quantity-prediction}}}                                                                                      \\
		$BERT+DExp     $                    & \textbf{0.7365}          & \textbf{0.3333}           & \textbf{1.4380}           & \textbf{109.9340}        \\
		\bottomrule
	\end{tabular}
	}
	\caption{Model Comparison}
	\label{table:quantity-prediction-1-model-ablation}
\end{subfigure}
\hfill
\vspace{0.2cm}
\begin{subfigure}[c]{\linewidth}
    \scalebox{1}{
	\begin{tabular}{l r r r r}
		\toprule
		                                               & \textbf{E-Acc$\uparrow$} & \textbf{LMAE$\downarrow$} & \textbf{MAPE$\downarrow$} & \textbf{MAE$\downarrow$} \\
		\hline
		$T_{Name} + T_{Type}$                          & 0.7086                   & 0.3881                    & 2.8679                    & 130.5220                 \\
		$T_{DescName} + T_{Type}$                      & 0.7245                   & 0.3560                    & 2.5356                    & 120.2433                 \\
		\hline
		$T_{DescName} + T_{Type} + Title $             & 0.7241                   & 0.3650                    & 2.3076                    & 126.5792                 \\
		$T_{DescName} + T_{Type} + OtherIngs $ & 0.7314                   & 0.3449                    & 1.8184                    & 117.8181                 \\
		$T_{DescName} + T_{Type} + Tags $              & 0.7306                   & 0.3423                    & 1.6525                     & 119.0621                  \\
		$T_{DescName} + T_{Type} + Servings $          & 0.7220                    & 0.3614                    & 1.9253                     & 127.7833                  \\
		\hline
		\multicolumn{1}{c}{\textbf{\modelname~for Task \ref{task:quantity-prediction}}}                   \\
		$T_{DescName} + T_{Type} + All $                 & \textbf{0.7365}          & \textbf{0.3333}           & \textbf{1.4380}           & \textbf{109.9340}        \\
		$T_{DescName} + T_{PredType} + All $          & \textbf{0.7339}          & \textbf{0.3353}           & \textbf{1.4412}           & \textbf{109.7673}        \\
		\bottomrule
	\end{tabular}
	}
	\caption{Input Ablation}
	\label{table:quantity-prediction-2-input-ablation}
\end{subfigure}
\caption{Ingredient Quantity Prediction}
\end{table}

\subsection{Experimental Results}

Table~\ref{table:measurement-type-prediction} shows the experimental results of the Ingredient Measurement Type Classification Task.
The results demonstrate that weight freezing in BERT (${BERT}^{freeze}$) showed relatively poor performance in classifying measurement types of target ingredients compared to W2V-based baselines.
In addition, training the weight parameters in BERT without using pre-trained weights ($BERT^{init}$) hinders the model's ability to distinguish the measurement types of ingredients compared to fine-tuning the weights ($BERT$).
Leveraging the \typetwo textual elements of recipes such as descriptive tags ($+ All$) did not benefit or rather incurred performance loss in model ablations using $BERT^{freeze}$ and $BERT^{init}$.
These results may seem to undermine the advantages of incorporating the recipe semantics.
However, our model performance results showed that not only does fine-tuning the BERT weights outperforms its related ablations but also offers improved synergy with the given recipe context represented as \typetwo textual elements.
Overall, \modelname~outperformed the baseline and its ablations in the Ingredient Measurement Type Classification task.

\begin{table*}[]
\renewcommand*{\arraystretch}{1.25}
\scalebox{0.6}{
\begin{tabular}{c|l|l|r|r|r|r|r}
\toprule
\multirow{2}{*}{\begin{tabular}{c} \textbf{Target}\\\textbf{Ingredient} \end{tabular}}  &
\multirow{2}{*}{\begin{tabular}{c} \textbf{Recipe Title} \end{tabular}}  &
\multirow{2}{*}{\begin{tabular}{c} \textbf{Descriptive Name} \end{tabular}}  &
\multicolumn{2}{c|}{\textbf{Actual}}     &
\multicolumn{2}{c|}{\textbf{Prediction}}     &
\multirow{2}{*}{\begin{tabular}{c} \textbf{APE}\\(\%) \end{tabular}}
\\
& &
& \multicolumn{2}{c|}{\textbf{Original $\rightarrow$ (g,ml)}}
& \multicolumn{2}{c|}{\textbf{Converted $\leftarrow$ (g,ml)}}
\\
\hline
\multirow{3}{*}{\begin{tabular}{l} Chicken \end{tabular}} & French Style Pate... &  chicken livers (chopped) & 8 ounce & 226.4g & $8\frac{3}{8}$ ounce & 238.39 g  & 5.3 \\
 &  Chinese Chicken Coleslaw Salad & skinless chicken breasts & 9 ounce & 254.70 g & $8\frac{3}{4}$ ounce & 248.05 g & 2.6 \\
 &  Tagine of Chicken and Chickpeas & boneless skinless chicken thighs & 1 lb & 453.59 & 1 lb & 452.75g & 0.2  \\
\hline
\multirow{3}{*}{\begin{tabular}{l} Beef\end{tabular}} &  Beef and Mushroom Lasagna & ground beef & 1 lb & 453.59 g & 1 lb & 448.45 g & 1.1 \\
 &   Roast Beef Sandwiches...  &  thinly sliced roast beef  & 16 ounce & 452.80 g & $16\frac{1}{4}$ ounce & 458.30 g & 1.2  \\
 &   Bacon-Chili Beef Stew  &  lean stewing beef & $1\frac{1}{2}$ lb & 680.39 g & $1\frac{1}{2}$ lb & 660.28 g & 3.0 \\
\hline
\hline
\multirow{3}{*}{\begin{tabular}{l} Water\end{tabular}}  & ...Chicken Machboos...           & rose water                      & 3  tbs & 44.37 ml & $2\frac{3}{4}$ tbs & 40.06 ml  & 9.7             \\
                        & El Gallo's Mexican Rice          & hot water                                & 2 cups      & 438.8  ml & 1 $\frac{7}{8}$ cups   & 473.18 ml & 7.3                       \\
                        & ...Chicken Ginseng Soup...       & water                                    & 10 cups     & 2365.9 ml & 10 $\frac{1}{8}$ cups & 2399.75 ml & 1.4                     \\
\hline
\multirow{3}{*}{\begin{tabular}{l} Pepper\end{tabular}} & French Roast Spice Rub           & ground black pepper             & $\frac{3}{4}$ ts & 3.7 ml & $\frac{2}{3}$ ts & 3.3 ml  & 10.7              \\
                        & ...Cream Cheese Canapes          & red bell pepper, finely chopped & $\frac{1}{4}$ cup & 59.15 ml & $\frac{1}{4}$ cup & 50.02 ml & 15.4                 \\
                        & Beef Wellington...               & diced red bell pepper           & 1 cup & 236.59 ml & 1 cup & 239.42 ml & 1.2                         \\

\bottomrule
\end{tabular}
}
\caption{
Prediction examples for target ingredients \textit{chicken}, \textit{beef}, \textit{water}, \textit{pepper}. Recipe attributes \textit{Target Ingredient, Recipe Title, Descriptive Name} and \textit{Original} in \text{Actual} are from the actual recipes in $\test$. Columns with \textit{(ml)} and \textit{(g)} are normalization units for \textit{Volume} and \textit{Weight}-based measurements respectively. All predicted quantities are converted back to their original units. APE refers to the Absolute Percentage error between normalized quantities.}
\label{table:inference-example}
\end{table*}

Similar comparison results for our next Ingredient Measurement Unit Classification task were also shown in Table~\ref{table:measurement-unit-prediction} which further justifies our choice of fine-tuning the pre-trained weights of BERT and using all \typetwo textual elements.
We remark that all experiments except ${T}_{DescName}+T_{PredType}+All$ used the actual ground truth labels of target ingredient measurement.

In Table~\ref{table:quantity-prediction-1-model-ablation}, we present the experimental results for our Ingredient Quantity Regression Task. 
Our proposed model not only outperformed its encoder and prediction layer ablations across all four evaluation metrics, but also demonstrated the importance of the DExp component in accurately predicting ingredient quantities. 
By leveraging the BERT model along with the DExp component, our model was able to capture the nuances of numeral elements and generate robust predictions that outperformed not only simpler point-wise estimation methods ($L1Loss$) but also a simpler version of likelihood-based training selection ($LogLP$). 
We conclude that our model's ability to capture the overall distribution of scales leads to more accurate and robust predictions.

Finally, Table~\ref{table:quantity-prediction-2-input-ablation} shows additional ablation results on modifying the input query composition for the Ingredient Quantity Regression Task.
Comparing input ablation $T_{Name}+T_{DescType}$ with $T_{Name}+T_{Type}$, the descriptive words surrounding the target ingredient name seemed to help \modelname~make more accurate predictions on ingredient quantities.
While using the target ingredient's descriptive name and measurement type, adding only other ingredients ($T_{Name}+T_{DescType}+OtherIngs$) or tags ($T_{Name}+T_{DescType}+Tags$) achieved comparable performance to \modelname~in all 4 evaluation metrics.

\textbf{Investigation on ablation results.} Overall, the experimental results for all three tasks demonstrate \modelname~'s ability to predict ingredient quantity properly through recipe comprehension and strengthen our rationale for its model design choices in implementation.
Our experimental results across all three tasks demonstrate the efficacy of our~\modelname~in accurately predicting ingredient quantities through recipe comprehension. 
Our model achieves this by leveraging the power of its individual components. Specifically, our encoder component utilizes the BERT model, which is shown to be superior to W2V-based model selection in capturing textual recipe semantics, as we have clarified in Table~\ref{table:quantity-prediction-1-model-ablation}. 

Additionally, our use of multi-modal likelihood in the model's regression layer allows for the accurate prediction of various scales of quantities ($DExp$), outperforming both point-wise estimators ($MLP+L1Loss$) and single-modal likelihood models ($LogLP$). 
These results reinforce the rationale for our model design choices in implementation and underscore the potential value of our approach for practical applications in recipe prediction.
We can infer \modelname~learns numerical cooking knowledge from structured multifaceted recipe contexts using the transferred knowledge from BERT and the Transformer model's ability compared to baseline approaches.

\section{Qualitative Analysis}
\subsection{Examples of Ingredient Quantity Prediction}
We performed qualitative analysis to investigate~\modelname's ability to understand recipe context and predict reasonable quantities of a target ingredient.
We firstly pivoted the target ingredients (e.g., water and pepper) and selected data instances from $\test$ that have different input compositions including their descriptive names.
We then used our trained \modelname~to predict the quantities annotated with units for each instance and compared them with actual numeral quantities used in recipes.

Table~\ref{table:inference-example} shows the predicted quantities of target ingredients with different recipe contexts.
We examined~\modelname's quantity and unit predictions based on different recipe contexts.

The first six rows in Table~\ref{table:inference-example} show the \modelname's predictions for pivoted ingredients \textbf{chicken} and \textbf{beef}.
Different parts (thighs, breasts, livers) or processing methods (ground, roast) can diversify the descriptive features of these meat-based ingredients.
All six data instances were predicted with less than 6\% error by \modelname.

The next three rows in Table~\ref{table:inference-example} show the \modelname's predictions for pivoted ingredient \textbf{water}.
While \textbf{water} can be used in various recipes, have different characteristics (hot, rose), and wide range of actual quantities normalized in milliliters (40.06, 473.18, 2399.75), all three data instances were predicted with less than 10\% error by \modelname.
We can infer that the recipe semantics such as title helped \modelname~capture varying magnitudes from $10^1$ to $10^3$ to make accurate quantity predictions for \textbf{water}.

The last three rows in Table~\ref{table:inference-example} show the \modelname's predictions for pivoted ingredient \textbf{pepper}.
Different from \textbf{water}, \textbf{pepper} can be either a plant-based (e.g., bell peppers) or spice-based (e.g., black pepper) ingredient which is actually different ingredients.
For the three data instances,~\modelname~showed modest performance in predicting quantities despite the ingredients having different processing methods (diced, ground), different physical states (dry, powder), and varying magnitudes in their actual quantities.
Interestingly, for the \textit{Beef Wellington} recipe, ~\modelname~seemed to understand that cut slices of pepper can be measured with volume-based units though most dry ingredient quantities are commonly measured with weight-based ones.

\subsection{Demo Program for Ingredient Quantity Prediction}
\begin{figure}
\centering
	\frame{\includegraphics[width=0.5\textwidth]{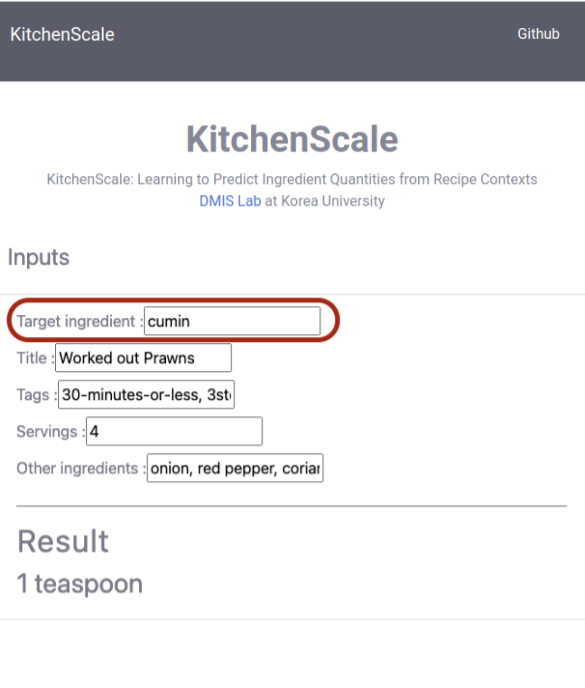}}
	\caption{Web Demo Program for Ingredient Quantity Prediction}
	\label{fig:demo}
\end{figure}

Figure~\ref{fig:demo} shows a snapshot of \modelname~recommending the adequate quantity for the target ingredient (cumin) given its recipe context. In this web-based application, \modelname~is provided with a target ingredient (cumin) and other recipe contextual features such as title (Worked out Prawns) and other ingredients (onion, red pepper, etc.) provided by the user as input. The trained prediction modules in \modelname~predict the measurement unit (teaspoon) and the normalized value of the target ingredient (cumin). Lastly, the normalized value (4.929) is converted to an actual quantity aligned with the predicted unit which is the final output for the user (1 teaspoon). The demo is available at http://kitchenscale.korea.ac.kr.

\section{Conclusion}
We devised \modelname~to predict the quantity and units of recipes via capturing recipe context through three consecutive prediction tasks. 
Our proposed model aims to assist users in determining precise ingredient quantities based on the textual description of a recipe. 
To achieve this, we leveraged the natural language processing capabilities of the BERT language model, which has been shown to be effective in capturing the semantics of text. 
By fine-tuning our model on a large corpus of recipe data, we were able to develop a tool that can generate accurate ingredient quantities based on a given recipe description.

The model first predicts the measurement type using the target ingredient and the recipe context as a query, and the quantity and the unit.
After the three prediction tasks, the final convert layer summarizes the outputs into a user-friendly format similar to existing recipes.
We implemented \modelname~using BERT as a recipe context encoder with a separator strategy to merge multi-type inputs and DExp method known as robust for a regression task in which the variance of output scale is high.
Experimental results show our approaches capture scales of ingredient quantities than other baseline approaches.
Moreover, the qualitative analysis provides insights into the final outputs.

Our empirical results show that our model selection, along with three novel task formulations, can lead to improved recommendation performance across various evaluation metrics on each task. 
This suggests that our model has the potential to be a valuable tool in the field of recipe recommendation systems.
We publish our codes with new datasets merging RecipeDB and Reciptor.

\section{Future work}
In the future, we expect our data-driven \modelname~to benefit from giving clues to make a numerically enhanced cooking assistant.
Since \modelname~is based on PLM, we would like to apply easily \modelname~to other downstream tasks like recipe generation and recipe retrieval, utilizing numerical food knowledge in~\modelname.
Also, we hope to integrate \modelname~with other food computing systems for food pairing, ingredient recommendation, and others.

Furthermore, we anticipate that \modelname~can be improved through personalization. 
Currently, the model cannot capture individual behaviors such as queries and reactions. 
However, we believe that incorporating personalized computing techniques~\citep{r2_sun2022eyes} will enhance the user experience through adjusting recommendations~\citep{r2_li2021hyperbolic} by providing feedback, rectifying errors, and continually updating the knowledge base of \modelname.

\section*{Funding}
This research was supported by the National Research Foundation of Korea (NRF-2020R1A2C3010638, NRF-2022R1F1A1069639), the MSIT(Ministry of Science and ICT), Korea, under the ICT Creative Consilience program(IITP-2023-2020-0-01819) supervised by the IITP(Institute for Information \& communications Technology Planning \& Evaluation) and Sony AI (https://ai.sony).

\section*{Acknowledgement}
Our work is part of a collaboration with Sony AI and their Gastronomy Flagship  Project, and our aim is to deploy \modelname~in food-related applications to help chefs and home cooks create delicious, healthy, and sustainable dishes by assisting them in determining appropriate ingredient amounts.
We anticipate that \modelname~will have a significant impact on the culinary domain and accelerate development in the food industry.

I would like to express my sincere gratitude to Donghyeon Kim and Sangrak Lim for their invaluable assistance throughout the development of this paper. 
Their insightful discussions and guidance greatly contributed to shaping my ideas and improving the quality of the work. 
I am truly grateful for their support.

\bibliographystyle{elsarticle-harv}
\bibliography{base}

\end{document}

%% file: math_commands.tex
\def\eqref#1{equation~\ref{#1}}

\def\floor#1{\lfloor #1 \rfloor}
\def\1{\bm{1}}

\DeclareMathAlphabet{\mathsfit}{\encodingdefault}{\sfdefault}{m}{sl}
\SetMathAlphabet{\mathsfit}{bold}{\encodingdefault}{\sfdefault}{bx}{n}

\def\num{y}
\newcommand{\train}{\mathcal{D_{\mathrm{train}}}}
\newcommand{\valid}{\mathcal{D_{\mathrm{valid}}}}
\newcommand{\test}{\mathcal{D_{\mathrm{test}}}}

\def\pred{\hat{y}}

\newtheorem{problem}{Problem}

\newtheorem{task}{Task}